\documentclass[10pt, journal, letterpaper]{IEEEtran}

\usepackage[hyphens]{url}
\usepackage{float}
\usepackage{placeins}
\usepackage[bottom]{footmisc}

\usepackage{color}

\usepackage{cite}


\usepackage{verbatim}
\usepackage{cite}
\usepackage{graphicx}
\usepackage[caption=false]{subfig}
\usepackage{epsfig}

\usepackage[cmex10]{amsmath}
\usepackage{amsthm}
\usepackage{thmtools,thm-restate}
\usepackage{amssymb}
\usepackage{mathrsfs}
\usepackage{algorithm}
\usepackage{algpseudocode}

\usepackage{array}
\usepackage{multirow}
\usepackage{makecell}
\setcellgapes{1.28pt}

\usepackage{cleveref}
\crefformat{section}{\S#2#1#3}
\Crefformat{section}{\S#2#1#3}

\IEEEoverridecommandlockouts
\AtBeginDocument{
	\abovedisplayskip=6pt plus 3pt minus 6pt 
	\abovedisplayshortskip=4pt plus 2pt minus 4pt
	\belowdisplayskip=6pt plus 3pt minus 6pt 
	\belowdisplayshortskip=4pt plus 2pt minus 4pt
}
\usepackage{makecell}

\setcellgapes{1.28pt}

\begin{document}
%
\title{Real-Time Pedestrian Detection on IoT Edge Devices: A Lightweight Deep Learning Approach} 

\author{
\IEEEauthorblockN{Muhammad Dany Alfikri \IEEEmembership{Student Member, ~IEEE}} \\
\IEEEauthorblockA{\textit{Department of Computer Science and Information Engineering} \\
\textit{National Taiwan University of Science and Technology}\\
m10915814@mail.ntust.edu.tw} \\
\and
\IEEEauthorblockN{Rafael Kaliski \IEEEmembership{Senior Member, IEEE}} \\
\IEEEauthorblockA{\textit{Department of Computer Science and Engineering} \\
\textit{National Sun Yat-sen University}\\
rkaliski@ieee.org$^{1}$}
\thanks{$^{1}$corresponding author}
\thanks{Rafael Kaliski is thankful for funding from Taiwan’s 
    National Science and Technology Council (NSTC), grants: NSTC 108-2218-E-011-036-MY3 and NSTC 112-2221-E-110-023-}%
}

\maketitle 

\begin{abstract}
Artificial intelligence (AI) has become integral to our everyday lives. Computer vision has advanced to the point where it can play the safety critical role of detecting pedestrians at road intersections in intelligent transportation systems and alert vehicular traffic as to potential collisions. Centralized computing analyzes camera feeds and generates alerts for nearby vehicles. However, real-time applications face challenges such as latency, limited data transfer speeds, and the risk of life loss. Edge servers offer a potential solution for real-time applications, providing localized computing and storage resources and lower response times. 
Unfortunately, edge servers have limited processing power. Lightweight deep learning (DL) techniques enable edge servers to utilize compressed deep neural network (DNN) models. 

The research explores implementing a lightweight DL model on Artificial Intelligence of Things (AIoT) edge devices. An optimized You Only Look Once (YOLO) based DL model is deployed for real-time pedestrian detection, with detection events transmitted to the edge server using the Message Queuing Telemetry Transport (MQTT) protocol. The simulation results demonstrate that the optimized YOLO model can achieve real-time pedestrian detection, with a fast inference speed of 147 milliseconds, a frame rate of 2.3 frames per second, and an accuracy of 78\%, representing significant improvements over baseline models.
\end{abstract} 

\begin{IEEEkeywords}
\textbf{Lightweight deep learning model}, \textbf{Pedestrian detection}
\end{IEEEkeywords}

\IEEEpeerreviewmaketitle

\section{Introduction}
\label{cha:intro} 

    AI has become indispensable in everyday life due to its ability to learn and adapt based on empirical data. It has found application in various fields such as medicine, governance, finance, communications, and transportation.
    Machine Learning (ML) offers a means of gaining insight into and addressing the ever more complex environments and demands requested of AI. However, ML typically requires hand-tuned features to perform well. By comparison, deep learning (DL) can automatically extract latent features using a general-purpose learning procedure \cite{lecun2015deep}. 

	DL has several applications, one being Computer Vision, where the DL model extracts high-level information and features similar to a human-level visual understanding based on visual input. There are areas related to computer vision, such as object detection, image segmentation, video tracking, object recognition, and motion estimation. Object detection and video tracking can work in tandem to perform pedestrian detection.  
	Pedestrian detection is a critical task for intelligent transport systems and autonomous vehicles. As such, it has gained interest and investment from 
    corporations and governments \cite{tokuyama1996intelligent}\cite{lim2012intelligent}\cite{njord2006safety}.  
	
	Based on the traffic accident data\cite{accident_data} from the Republic of China's Ministry of Transportation and Communication, in 2021, there were 16,649 instances of accidents involving pedestrians in Taiwan. In those accidents, approximately 90\% of the people involved sustained bodily harm, and 410 people lost their lives. In densely populated countries, this data is worrying. While many factors are involved in accidents, one of the most common is human error. Human error occurs due to pedestrian and driver errors, such as being distracted. Some examples of human errors involve pedestrians who ignore their surroundings while walking, look at their phones, or illegally cross the road. The driver who uses their phone while driving is the most common example of human error involving drivers. While enacting laws can help reduce human errors, intelligent transportation systems must make roads safer for pedestrians and drivers. 

    Intelligent transportation systems enable the use of video and photos for object recognition.       Multiple traffic cameras are deployed at signalized intersections and can detect pedestrians accurately. Then, cloud computing processes the camera feed to detect pedestrians and generate warnings for nearby vehicles around the intersections. However, there are several issues related to cloud computing. Some of them include latency, limited data transfer speed, and the possibility of packet loss, which makes cloud-based computing unsuitable for real-time applications \cite{dimitrakopoulos2010intelligent}. The other issue is that to receive warnings, the vehicle must have an onboard unit (OBU), which consists of multiple sensors that send the driver information situation-related data. OBUs communicate with surrounding vehicles and network infrastructure through dedicated short-range communication (DSRC) \cite{sterzbach1996mobile}. 

    One method to address network latency is to place edge servers in the radio access network. This method transfers computing and storage from the cloud to the edge to reduce latency \cite{WANG2019160}. The network can better handle response time-critical situations by using the edge servers. Due to their capabilities, Edge Servers often perform various AI tasks, such as object detection \cite{gao2020salient}, anomaly detection \cite{8844663}, and reinforcement learning \cite{9394240}.

    While edge servers have many advantages, they also have many disadvantages. Compared to cloud computing servers, edge servers have lower computing power due to their compact form factor. Edge servers can overcome this limitation by using methods and models that do not require high processing power. One such model is lightweight DL, wherein a compressed DNN model with smaller, more efficient models with similar performance metrics to the original model executes on edge devices\cite{lightweightdl}.    

\subsection{Research Contributions}
    In this research, we design an edge-device-compatible computer vision-based lightweight DL model for pedestrian detection, employing a novel process for detecting and localizing an object. We then train and evaluate the DL model, optimize it to become lightweight, and compare the results to other baseline models. Finally, we deploy the model on an edge device (Nvidia Jetson Nano) to test the model's capabilities in a real-time scenario.

    The contributions of this research are as follows:
    \begin{itemize}
        \item We design a lightweight DL model that can be deployed on an inexpensive edge device to detect pedestrian traffic in real-time.
        \item We develop an AI on edge system that can operate in a low bandwidth environment by transmitting the essential information of interest.
        \item We compare our model to other popular non-YOLO models and demonstrate that other models require at least $3\times$ the computing power for over $2\times$ the memory for similar accuracy metrics.
    \end{itemize}

\subsection{Research Organization}
In \cref{sec:literaturereview}, we present related works. Then, in  \cref{sec:sysmodel}, the system model and proposed methods are presented. The simulation setup and topology are discussed in \cref{sec:sim_setup}. Then, the simulation results and analysis are presented in \cref{sec:simresults}. Finally, in \cref{sec:conclusion}, the research conclusions and future research directions are presented.

\color{black}

\section{Related Works}
\label{sec:literaturereview}
Pedestrian detection is a computer vision problem that involves recognizing the presence of pedestrians in image or video sequences while ignoring other objects. It has found application in autonomous driving \cite{yang2018real} \cite{cui2008pedestrian} \cite{PBP}, monitoring\cite{wang2009pedestrian}\cite{shao2021real}, and human-computer interfaces\cite{brunetti2018computer}. 
Pedestrian detection algorithms can be categorized into two main approaches: handcrafted features, such as extracting channel features that consist of different color \cite{dollar2009integral} and gradient channels \cite{costea2016semantic}, and DL, such as using convolutional neural networks (CNNs) \cite{girshick2015fast}\cite{ren2015faster}. Handcrafted feature-based methods use features that capture pedestrians' visual characteristics, such as shape, texture, and color. These features are then fed into a classifier, such as a support vector machine (SVM), to distinguish between pedestrian and non-pedestrian regions in the image. On the other hand, DL-based methods use CNNs to automatically learn to discriminate features from the raw pixel values of the image. These features are then used to classify each region in the image as pedestrian or non-pedestrian.
Both algorithms typically consist of two main stages: candidate generation and classification. Candidate generation involves generating a set of regions in the image that may contain pedestrians. Sliding windows or region proposal methods, such as selective search or region-based CNNs (R-CNNs) \cite{uijlings2013selective}, assist in pedestrian detection. Classification involves determining whether each candidate region contains pedestrians or not \cite{girshick2015fast}, using the classifier trained on handcrafted or DL-based features.

\subsection{Handcrafted Features based Pedestrian Detection}

In the early development of pedestrian detection, various handcrafted features were used. Handcrafted features consist of variations of color, texture, or edge \cite{benenson2014ten}. The most used handcrafted feature in pedestrian detection is the histogram of gradients (HOG). Its popularity is mainly due to its ability to extract a specific shape from the background images. The algorithm used for HOG can obtain the shape by sliding a detection window over grids of overlapping cells to form a block called a HOG feature vector \cite{hog}. 
The handcrafted features are usually fed into ML algorithms like SVM \cite{bertozzi2007pedestriansvm}, boosting algorithms \cite{daniel2017boostin}, and others. In  \cite{he2008scale} \cite{zhou2009human}\cite{zhang2010pedestrian}, various methods of HOG feature extraction are used together with SVMs as a classifier. While in \cite{cui2008pedestrian}, HOG features are used as training data for the ML model that combines the AdaBoost and SVM classifiers.

Due to the growing interest in real-time applications and the increased capability of computer processing power, the handcrafted feature methods are no longer practical due to their inflexibility owing to their requirement to search the entire image to detect objects at different locations and scales, making the model slow and impractical to use in a scenario where time is critical. New methods relied on DL, which is believed to have the ability to generate features independently and is, therefore, not reliant on handcrafted feature generation. As such, DL models are considered faster and more flexible.

\subsection{Deep Learning based pedestrian detection}

Following the development of AI, DL models were developed for use in various computer vision tasks, such as image classification\cite{hinton2012improving} \cite{chan2015pcanet}, semantic segmentation \cite{visin2016reseg}\cite{badrinarayanan2017segnet}, and object detection \cite{girshick2015fast}\cite{ren2015faster}. Based on the DL model's ability to perform various computer vision tasks, models were developed to tackle the pedestrian detection problem in recent years. DL-based pedestrian detection can be divided into two types: hybrid-based pedestrian detection and fully DL-based pedestrian detection.

In hybrid-based pedestrian detection, DL models use CNNs to extract features from the input images using a predetermined step direction. Following the step, the CNN yields a detection window containing the features classified as a positive class (contains an object) or a negative class. The extracted features are then used to train a shallow classifier, such as an SVM or a boosting algorithm. In \cite{hu2017pushing}, the author utilizes an ensemble of boosting algorithms trained on the features collected from the CNN. While in \cite{zhang2016faster} and \cite{tesema2020hybrid}, region proposal network\footnote{A region proposal network is a fully convolutional network that simultaneously predicts object bounds and objectness scores at each position.} was proposed in \cite{ren2015faster} is used as an initial detector, the results of which refined via a shallow network to improve detection results.

In fully DL-based pedestrian detection, DL models are trained end-to-end to learn how to perform feature extraction and how to detect and classify the results. The step usually includes training the backbone network, which is usually a CNN using backpropagation methods that work by propagating the error from the output layer back through the network to adjust the weights of each neuron in each layer. The error is calculated as the difference between predicted and actual output. Afterward, the error is used to adjust the weights in each layer using gradient descent; therefore, the CNN would be able to extract features automatically.
The next step is to generate proposals, explained as candidate regions in the image that may contain objects, in this case, pedestrians. This process can use sliding window techniques \cite{girshick2015fast} or a region proposal network\cite{zhang2016faster}. Then, a CNN will classify each proposal from the previous step into a different class, usually one containing objects. The final step is to apply post-processing techniques such as non-maximum suppression \cite{neubeck2006efficientnms}, sorting the detected bounding boxes by their confidence scores, representing the likelihood that the box contains an object of interest. Then, starting with the highest-scoring box, it is compared to all the other boxes that have not yet been suppressed. If the overlap between the two boxes,  measured by their intersection over union, is above a certain threshold, then the box with the lower score is suppressed. We repeat this process for all remaining boxes until no more boxes can be suppressed. The threshold is often set at 0.5, so if two bounding boxes overlap by more than $50\%$, we consider them duplicates and suppress the one with a lower confidence score.

While hybrid-based methods can achieve good detection performance, they are typically slower than DL-based methods due to their reliance on handcrafted features and separate proposal generation and classification steps. Additionally, hybrid-based methods require more manual parameter tuning than end-to-end trained DL models.
In a DL-based method, the CNN is trained end-to-end to learn the feature extraction and classification steps. The CNN generates proposals directly from the input images and then classifies each as containing a pedestrian or not using the same CNN. This approach eliminates the need for handcrafted features and allows faster and more accurate detection performance. Therefore, it is more suitable for real-time use.

\section{System Model and Proposed Methods}
\label{sec:sysmodel}

\begin{figure}[t]
\centering
\includegraphics[width=0.7\columnwidth,clip,trim=150 15 15 50]{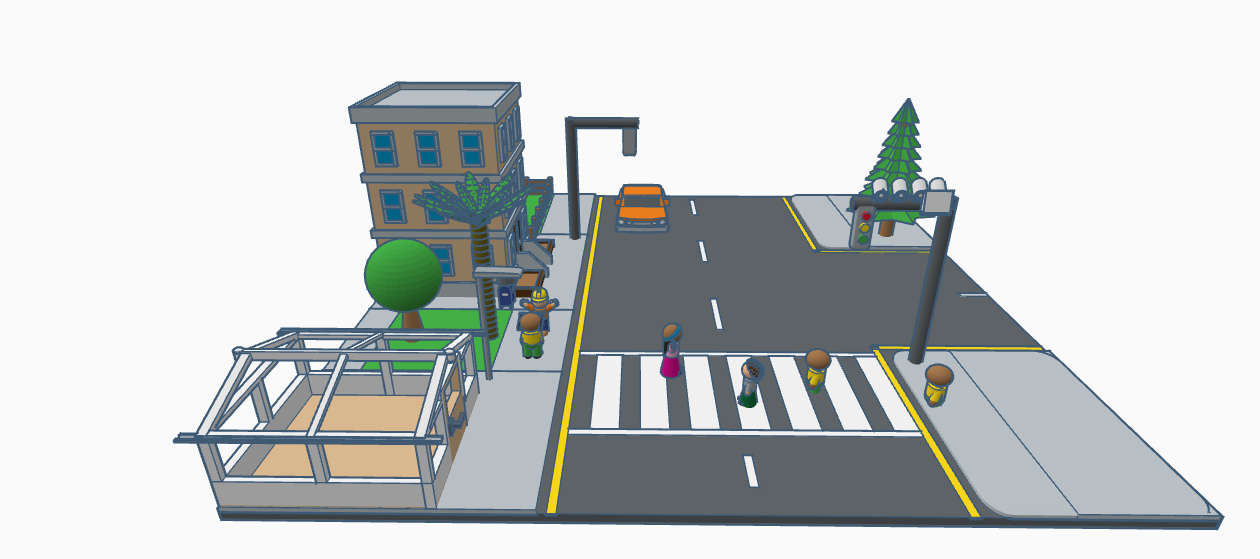}
\caption{Traffic light and intersection system model}
\label{fig:AI_Cam_Sys_Model}
\end{figure}
Our AI camera system is deployed on the traffic light infrastructure, as shown in Fig. \ref{fig:AI_Cam_Sys_Model}.  The AI camera captures pedestrians crossing the road. After detecting a pedestrian, the AI camera system forwards the captured images to the integrated edge devices for processing. The edge devices analyze the images and generate warnings or alerts for nearby drivers in case of any extraordinary occurrences or potential safety hazards. Due to the critical nature of these warnings for public safety, they are given high priority within the network infrastructure, thus ensuring that the alerts are promptly communicated to drivers, allowing them to take necessary precautions and respond accordingly.

\begin{figure}[t]
\centering
\includegraphics[width=0.5\columnwidth]{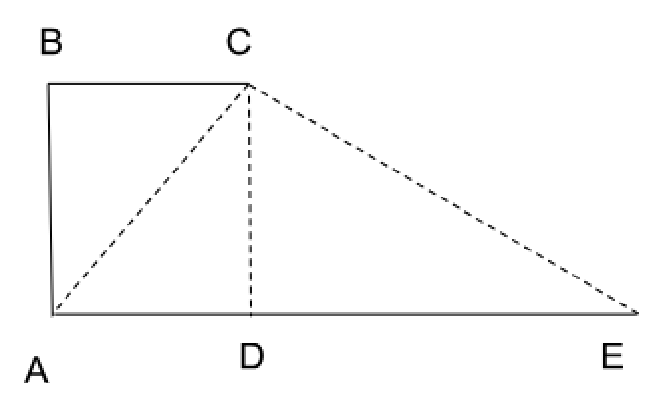}
\caption{Geometric diagram of camera deployment}
\label{fig:Geo_Traffic_Light}
\end{figure}

In the deployment, we assume the camera is mounted on the traffic light. Based on this assumption, we can estimate the camera's location based on the traffic light's height and width. We can estimate the camera's location per Fig. \ref{fig:Geo_Traffic_Light}.
For example, assuming that the height of the traffic light (AB) is 6 meters,  the width (BC) is 3 meters,  the length of the two road lanes (AE) is 9.2 meters, and the distance between the camera and the opposite road lane is 6.2 meters. Therefore, the distance from the side where the traffic light is mounted (AC) is 6.7 meters, while the distance from the opposite side of the crosswalk is 8.63 meters.

\begin{figure}[t]
\centering
\includegraphics[width=0.7\columnwidth]{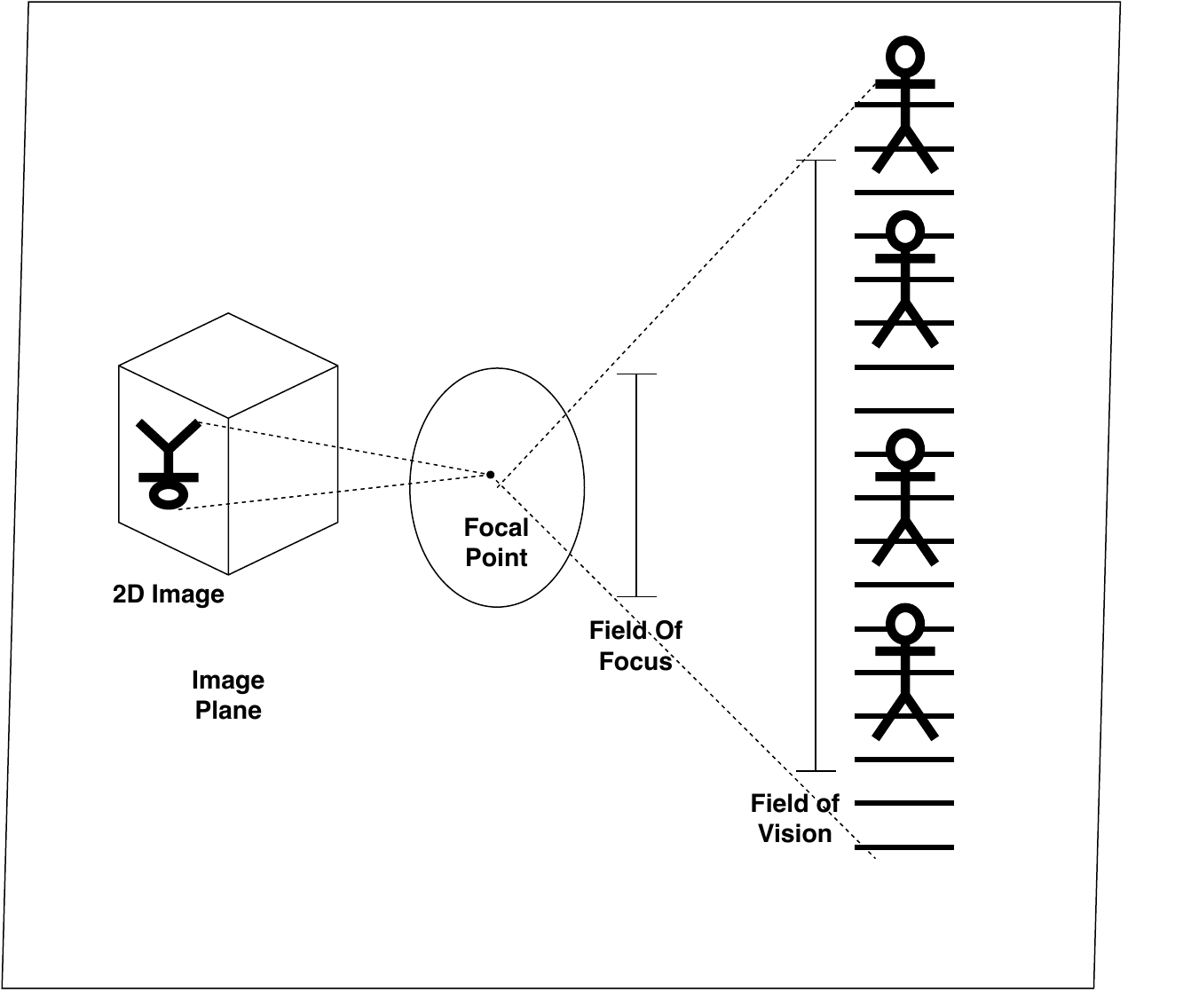}
\caption{Camera placement diagram}
\label{fig:Cam_Placement}
\end{figure}

Putting the camera on top of the traffic light provides the optimal view to detect pedestrians crossing the other side of the road. Fig. \ref{fig:Cam_Placement} details the detection area.
By mounting the camera on top of the traffic poles, the camera would have the optimal field of view (FoV) to capture data for the model, resulting in a larger and broader FoV. A greater FoV can capture more contextual information, potentially assisting the object detection algorithm to understand the scene better and make more accurate predictions. It can be helpful when objects of interest interact with others or the environment. A larger FoV can help the model understand the spatial relationships between objects, which is especially important when objects may occlude each other or when the scene requires more context for accurate detection.

\subsection{Proposed pedestrian Model for edge computing}

The most important component for real-time pedestrian detection is the DL model. In order to be deployed in real-time, DL must have a fast detection time and use minimal computational power to fulfill this requirement. DL models may be divided into two categories: single-stage and two-stage detectors. Single-stage detectors directly predict object bounding boxes and class probabilities from a single pass through the network without requiring additional region proposal processing steps. They use dense sampling and predefined boxes/key points of various scales and aspect ratios to localize objects in a single shot. Examples of this category are YOLO \cite{redmon2018YOLOv3}, Single Shot Detector \cite{liu2016ssd}, and RetinaNet \cite{lin2017focalretinanet}. On the other hand, two-stage detectors have a separate module to generate region proposals in the first stage, followed by the classification and localization of objects in the second stage. These models try to find an arbitrary number of potential objects in an image during the first stage. In the second stage, the model attempts to classify and localize objects. Examples of this category are faster RCNN \cite{ren2015faster}, EfficientNet \cite{tan2019efficientnet}, and CenterNet \cite{duan2019centernet}. This research used a YOLOv3-based DL model due to its fast detection time and minimal computational power. We used transfer learning to train the model. The main reason for using this method is the DL model's ability to adapt to new data. Because the DL model has acquired information from the previous data, its knowledge can be leveraged to learn new information. This is due to the ability of a CNN to recognize patterns with similar connections or features. 

We implement the model explained in \cite{tiny-YOLO} to further optimize the model to fulfill the requirement. The author explained that by reducing the depth of the convolutional layer, the model would have a decreased running speed, but its accuracy would decrease. 

The DL model will then be deployed on AIoT devices at road intersections to monitor pedestrian crossings. The AIoT devices stream real-time detection results to the edge server via an MQTT broker. The edge server then processes the results based on specific events. If the event is detected, the edge server sends a notification to the nearby vehicles.  

\subsection{Deep Learning Model}

In this research, we will implement a two-stage detection model, with each stage handling a specific computer vision task. Stage 1 will deal with object detection, which is used to determine the pedestrian's coordinates relative to the input received from the camera. Then, it will pass into Stage 2, which deals with pose estimation to predict the pedestrian's direction. The more detailed system model is explained in the figure below.

In this research, we implement a lightweight DL model optimized for IoT devices. One critical component of this optimization is reducing the depth of the neural network without reducing its performance. While deeper neural networks generally produce better results, their main disadvantage is the resource requirement. A DNN requires many resources, memory requirements, and computing power, which may not be available on edge devices with limited memory and computing power. 

We based our model on YOLO-V3 Tiny \cite{tiny-YOLO}. It was discovered that adding a pooling layer and reducing the depth of the convolutional layer can still achieve a high accuracy yet a faster inference time. In our DL model, we use a CNN architecture\footnote{In the YOLOV3 model, a Darknet-53 architecture consisting of 53 convolutional layers and bypass links was used to extract features.} and 
MobileNet, wherein the architecture implements a technique called inverted residuals and linear bottlenecks. The technique involves changing the input into a low-dimensional representation and then expanding the input model into a high-dimensional representation after implementing depth-wise convolution. A benefit of this architecture is that features can be extracted without requiring a deep convolutional layer, i.e., the architecture is considered mobile-friendly and could be implemented on edge devices. The model has three parts: the backbone, neck, and head. The backbone is a neural network architecture used to extract features.


\subsection{Model Backbone}
 In this research, we implement MobileNetv2 as the backbone network. The main reasons we used MobileNetv2 are the inverted residual blocks and the linear bottleneck layers.
 The inverted residual block consists of three components:
 \begin{enumerate}
  \item Expansion: In this step, the number of channels $C$ of the input are expanded by a factor $t$. When $t > 1$, the number of channels in the intermediate feature maps $O$ increases per \eqref{eq:MobileNetv2_expansion}.
  \begin{equation}
    \label{eq:MobileNetv2_expansion}
    O = C * t    
  \end{equation}
  \item Depthwise Convolution: The depthwise separable convolution involves applying a depthwise convolution with a kernel size of $k x k$ expanded on the feature maps. Depthwise convolution applies a separate convolution filter to each channel of the input. The number of depthwise convolution parameters $D$ is per \eqref{eq:MobileNetv2_Depth_Params}. 
  \begin{equation}
    \label{eq:MobileNetv2_Depth_Params}
    D = k^2 \times C    
  \end{equation}
  \item Pointwise Convolution (Projection) uses a 1x1 kernel applied to the feature maps to project back to the desired number of channels. The number of pointwise convolution output channels $C$. 
  The Total Number of Parameters in the Inverted Residual Block $T_P$ is per \eqref{eq:MobileNetv2_IRF_total_params}.
  \begin{equation}
      \label{eq:MobileNetv2_IRF_total_params}
      T_P = C * t^2 + k^2 * C + C * C
  \end{equation}
\end{enumerate}
In MobileNetV2, the output of an inverted residual block is the sum of the input to the block and the output of the last bottleneck layer. This "shortcut connection" aids in the preservation of information across. The input to the block is specifically passed through a bottleneck layer, which reduces the number of channels in the feature map, followed by a series of depthwise separable convolutions, which act as a spatial filter to the feature map. The depthwise separable convolutions are then followed by another bottleneck layer that restores the number of channels to its original size. After that, the output of the last bottleneck layer is added to the input of the inverted residual block to produce its output. This output is then passed to the next layer in the neural network.

\subsection{Model Neck and Head}

You Only Look Once (YOLO) version 3 is a popular object detection model known for its real-time performance and efficiency. It performs detection in a single pass over the image via a single fully connected layer, enabling high frame rates and responsiveness. The model utilizes multiple output layers with different scales to detect objects of various sizes effectively. Despite not achieving the highest accuracy, YOLOv3 balances speed and accuracy, making it suitable for real-world applications. It is versatile and adaptable to different object detection tasks and custom datasets. As YOLOv3 is open-source and has an active community, it receives continuous improvements and support. 

\begin{figure}[t]
\centering
\includegraphics[width=\columnwidth]{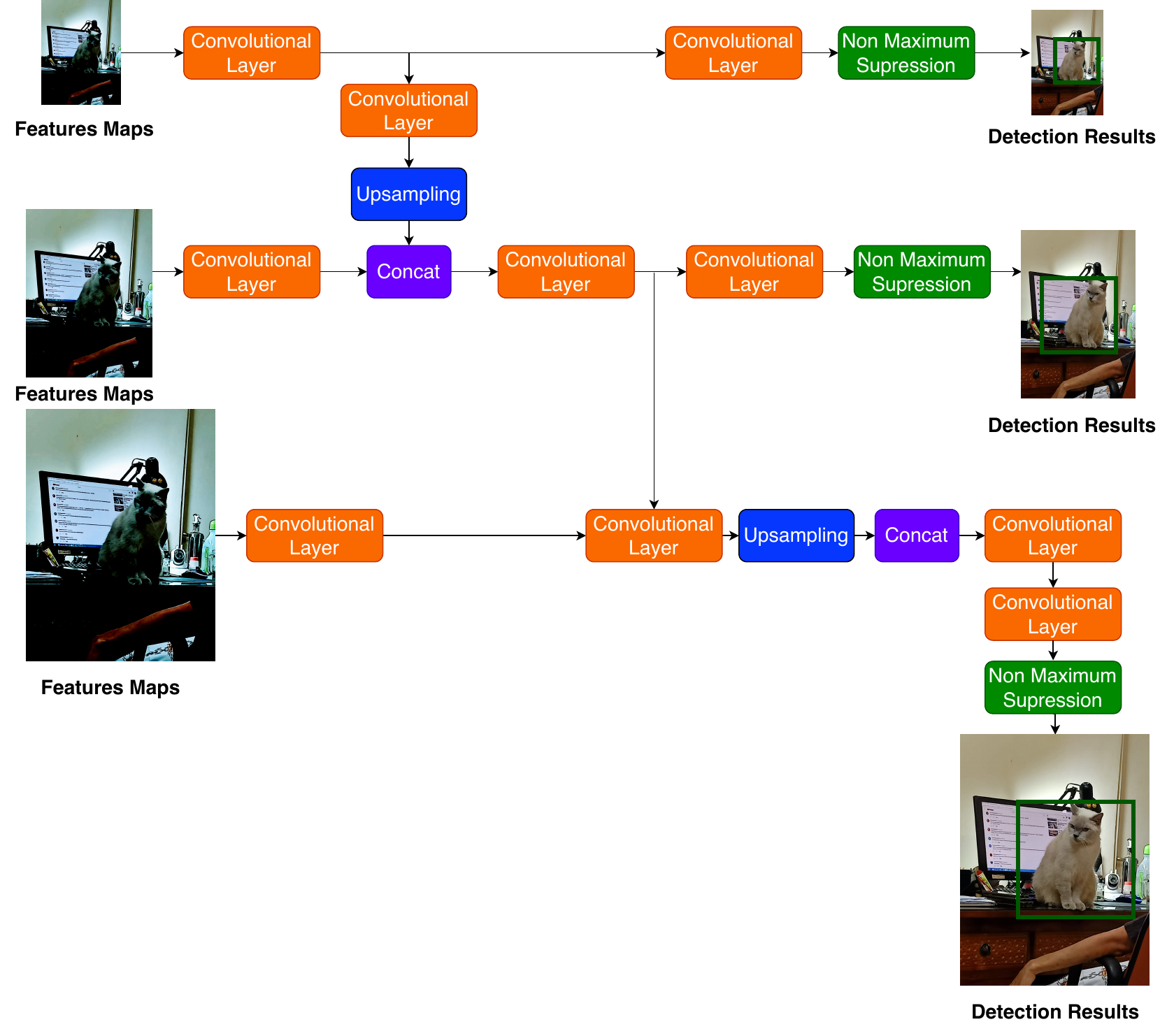}
\caption{YOLOv3-based deep learning model architecture}
\label{fig:YOLOv3_arch}
\end{figure}

YOLOv3 receives input with different scales of feature maps from the MobileNet backbone model. In YOLOv3, the network's neck utilizes three different scales of features for effective object detection: 
1) High-scale features originate from the output layer with the lowest stride (32), capturing fine details and being ideal for detecting smaller objects. 
2) Medium-scale features are derived from the output layer with a higher stride (16), offering a coarser spatial resolution and excelling at detecting medium-sized objects. 
3) Low-scale features come from the output layer with the highest stride (8), which provides the coarsest spatial resolution and specializes in detecting large objects. 
%
%
The corresponding detection heads predict bounding boxes, objectness scores, and class probabilities for small, medium, and large objects.
Combining these scales enhances YOLOv3's ability to handle objects of varying sizes efficiently and makes it a favored choice for real-time object detection tasks. The architecture of YOLOv3 is shown in Fig. \ref{fig:YOLOv3_arch}.

The convolutional layers in the neck of YOLOv3 extract visual features from the input feature maps. The convolutional layers vary in filter size and number to better capture features at different scales and complexities. The upsampling layer increases the spatial resolution of the feature map of a lower-scale detection layer to match the resolution of a higher-scale detection layer, ensuring concatenation compatibility. The concatenation layer combines feature maps from different scales along the channel dimension, allowing the model to fuse multi-scale information and more comprehensively represent the input image. This multi-scale feature fusion is critical for YOLOv3's ability to detect objects of varying sizes, making it robust for real-time applications.
The combined feature maps from different scales are further processed using additional convolutional layers after the concatenation operation in the neck of YOLOv3. These convolutional layers are critical in refining the multi-scale feature representation and extracting more abstract features for final object detection.  

\begin{figure}[t]
\centering
\includegraphics[width=\columnwidth]{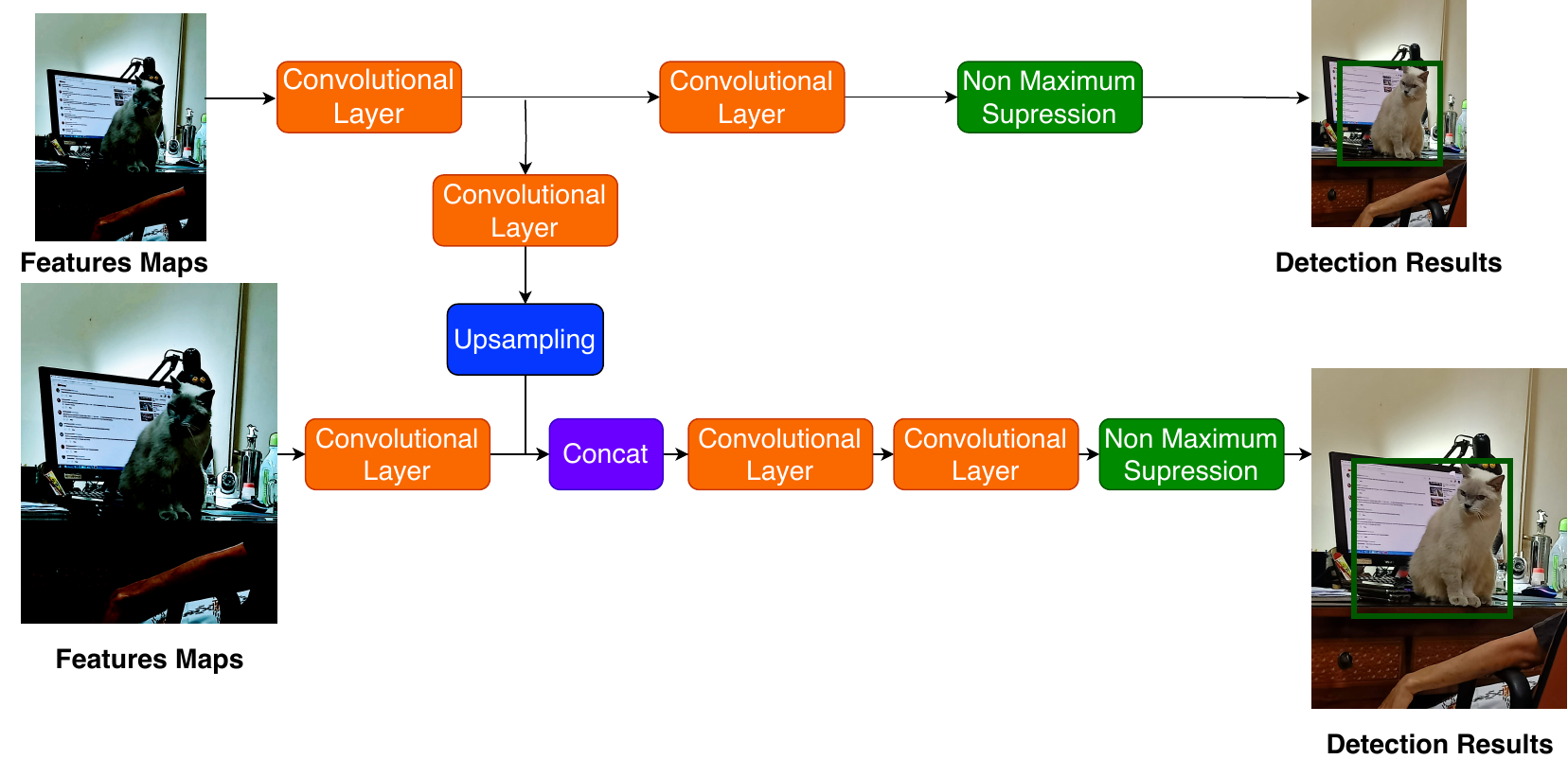}
\caption{Our optimized Tiny YOLO V3 model architecture}
\label{fig:Tiny_YOLOv3_arch}
\end{figure}

To optimize the model, we reduce the number of convolutional layers to make it more compact and efficient. 
However, balancing reducing model complexity and maintaining detection performance is critical. Removing too many convolutional layers may lose essential features and reduce the model's accuracy. Fig. \ref{fig:Tiny_YOLOv3_arch} shows our model's architecture.

In addition to reducing the convolutional layers, we use two detection scales instead of the three ones used in the YOLOv3 model. Tiny YOLOv3 can effectively handle objects of varying sizes in the image using these two different detection scales. Smaller objects in the image are detected using high-scale detection, while medium-scale objects are detected using low-scale detection. Tiny YOLOv3 performs high-scale (low-scale) detection on a feature map with a stride of 32 (16), i.e., each cell in this feature map corresponds to a region in the input image that has been reduced in width and height by a factor of 32 (16).  
Using this multi-scale approach, tiny YOLOv3 can achieve accurate and efficient object detection performance in real-time applications on resource-constrained devices. 

\subsection{Dataset}

We train our model using the Crowdhuman dataset \cite{shao2018Crowdhuman}. The Crowdhuman dataset comprises over 15,000 images with over 470,000 annotated human instances at different scales, viewpoints, and occlusion levels. The dataset is gathered from various locations, such as streets, shopping centers, and airports, and is captured using different types of cameras. It is usually used to train the model for tasks involving detecting people in dense crowds. The Crowdhuman dataset is large and rich and contains a high diversity of people in different occlusion scenarios. 


In order to prove that our model performs better than the available models, we compare widely used DL models pre-trained with the Crowdhuman dataset. The models that we compare with are Retinanet \cite{lin2017focalretinanet}, Faster RCNN\cite{zhang2016faster}, EfficientNet\cite{tan2019efficientnet}, and Single Shot Detector\cite{liu2016ssd}.

\subsection{MQTT protocol for Internet of Things}

In order to communicate effectively, IoT devices need reliable real-time protocols that operate in low-bandwidth unreliable networks. Initially developed by IBM \cite{hillar2017mqtt}, Message Queuing Telemetry Transport (MQTT) is a lightweight messaging protocol enabling IoT device communication. It is designed to be efficient and reliable, even in low-bandwidth and unreliable network environments. MQTT supports several Quality of Service (QoS) levels that ensure message delivery during network disruptions or failures.


MQTT operates on the publish-subscribe model wherein a client that publishes a message is decoupled from the other clients that receive the message in the publish-subscribe asynchronous pattern. Only one client receives the message, reducing the risk of leaked messages. 
A typical MQTT system 
consists of a publisher, a broker, and one or more subscribers \cite{hillar2017mqtt}. The device or client that sends messages to the broker is the publisher. The messages can contain various data, including sensor readings, device states, and commands, and are categorized by topic. 
The broker, an intermediary between publishers and subscribers, distributes messages to all connected topic subscribers. The broker is responsible for managing all the connections, handling the message queue, and ensuring that messages are only delivered to the intended recipients.
Subscribers, i.e., clients who receive broker messages, may elect to receive messages from one or more topics. 
After receiving a message, a subscriber can process or act on the message. Subscribers can also unsubscribe from topics if they elect to do so.

\section{Simulation Setup}
\label{sec:sim_setup}

Simulating pedestrian crossing scenarios using videos has several advantages. One of the main advantages is that it allows researchers to create controlled environments for testing and evaluating pedestrian models, algorithms, and systems. Simulations also allow for testing extreme scenarios that may not be safe or feasible to replicate in real-life testing. For example, simulating a pedestrian crossing in a dense traffic area at peak hours allows researchers to test the effectiveness of different traffic management strategies while minimizing pedestrian risks.

\begin{figure}[t]
\centering
\includegraphics[width=\columnwidth]{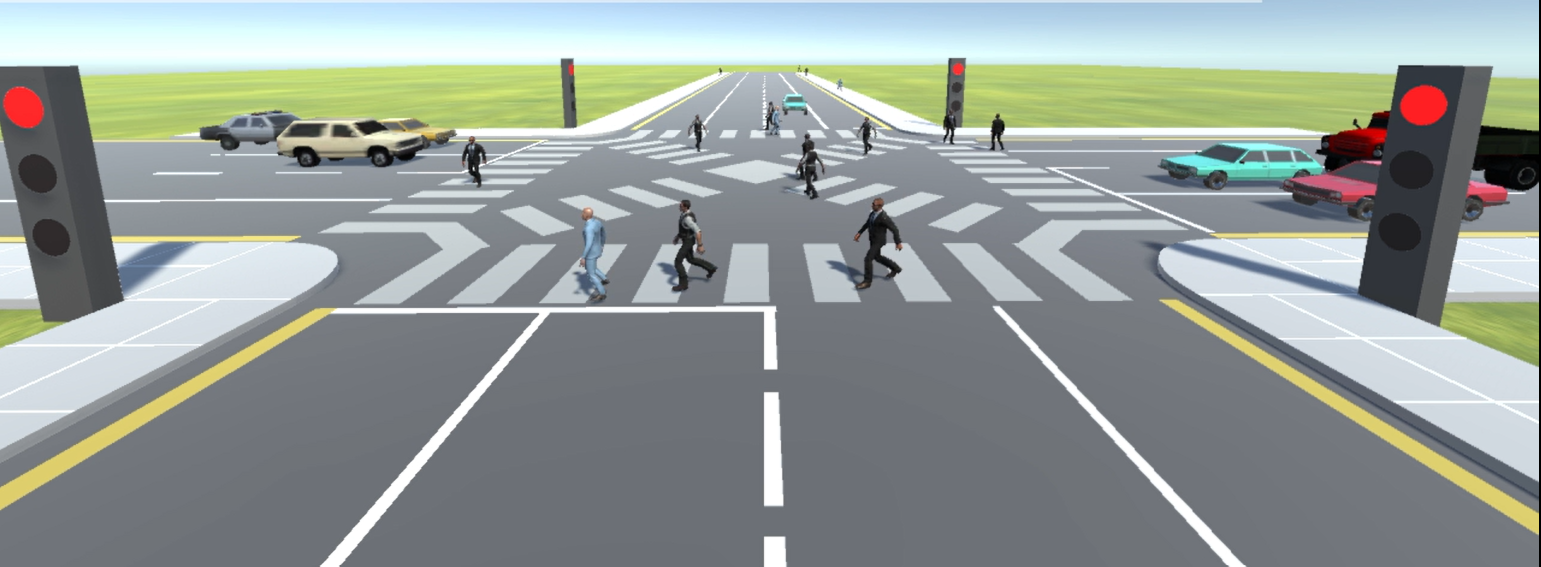}
\caption{Scenario of pedestrian crossing the street}
\label{fig:Unity_Sim}
\end{figure}

In order to replicate the pedestrian crossing the road scenario, in this research, we utilized Unity \cite{haas2014historyUnity} combined with the Traffic3D library. 
%
Unity is intended to be versatile and adaptable, allowing developers to create games and simulations of any genre or style. 
Traffic3D is a traffic simulation tool that offers a rich 3D environment for training intelligent traffic management agents. It is intended to create visually and physically intelligent traffic simulation models to test new technology for eventual deployment in the real world. 
Traffic3D allows for scene generation programmatically, even during runtime, and supports a plug-and-play architecture to ensure the stability and generalizability of agents trained with Traffic3D to variations of the environment. This research uses Traffic3D to simulate pedestrians crossing the street at an intersection. The scenario is shown in Fig. \ref{fig:Unity_Sim}.

\subsection{Deep Learning Model}
In this experiment, we will use MMDetection \cite{chen2019mmdetection}, an open-source framework based on Pytorch \cite{paszke2019pytorch} developed by the OpenMMLab team from Nanyang Technological University, to build custom DL models. 
Using the MMDetection framework, we train Retinanet \cite{lin2017focalretinanet}, Faster RCNN\cite{zhang2016faster}, EfficientNet\cite{tan2019efficientnet}, and Single Shot Detector\cite{liu2016ssd} as well as our optimized Tiny-YOLOv3 model on the Crowdhuman \cite{shao2018Crowdhuman} dataset using pre-trained weights 
to maintain fairness 
across each category of the protected features when evaluating the model.

Our model is then deployed on an inexpensive Nvidia Jetson Nano, i.e. a powerful single-board computer designed specifically for AI applications. 
A Jetson Nano is particularly well-suited for applications such as autonomous vehicles, robotics, and smart cameras, where high-performance processing and low power consumption are critical. The main Jetson Nano B01 parameters are detailed in table \ref{tbl:Jetson_Nano}. 
\begin{table}[b]
\caption{Simulation Specifications}
\label{tbl:Jetson_Nano}
\centering
\begin{tabular}{|p{5em}|p{22em}|}
\hline
\multicolumn{1}{|l|}{AIoT Device} & NVIDIA Jetson Nano B01 \\ \hline
\multicolumn{1}{|l|}{GPU}                            & {NVIDIA Maxwell architecture, 128 CUDA cores} \\ \hline
\multicolumn{1}{|l|}{CPU}                            & Quad-core ARM Cortex-A57 MPCore                            \\ \hline
\multicolumn{1}{|l|}{SW Config} & NVIDIA jetpack 4.6, Ubuntu 18.04, cuDNN 8.2.1, Cuda 10.2 
\\ \hline
\multicolumn{1}{|l|}{Memory}                         & 4 GB 64-bit LPDDR4                                                  \\ \hline
\multicolumn{1}{|l|}{Storage}                        & MicroSDXC 128GB A2                                        
\\ \hline
\multicolumn{1}{|l|}{Sim colorspace} & RGB \\ \hline
\multicolumn{1}{|l|}{Sim duration} & $40$ seconds \\ \hline
\multicolumn{1}{|l|}{Sim frame rate} & $30$ fps \\ \hline 
\multicolumn{1}{|l|}{Sim resolution} & $1280 \times 720$ \\ \hline
\multicolumn{1}{|l|}{Connectivity} & Wi-Fi (Intel AC8265) \\ \hline 
\multicolumn{1}{|l|}{MQTT Broker} & {Eclipse Mosquitto} \\
\hline
\multicolumn{1}{|l|}{MQTT Client} & {Paho MQTT} \\
\hline
\multicolumn{1}{|l|}{MQTT QoS} & {QoS 0} \\
\hline
\end{tabular}

\end{table}

\subsection{MQTT Broker Setup}
In this research, we use Eclipse Mosquito\cite{light2017mosquitto}, a small, lightweight MQTT server, as the MQTT Broker. 
%
In our testbed, we use a 
personal computer as the MQTT broker is linked to the Jetson Nano via a wireless network. 
%
The MQTT broker controls the message flow and ensures that publisher messages reach their intended subscribers.
The wireless network acts as a conduit for communication between the MQTT broker and the MQTT subscriber, both connected to the same wireless network. 

The MQTT subscriber, a device or an application, indicates a desire to be interested in specific topics and connects to the MQTT broker via the wireless network. 
This configuration simulates a wireless network environment similar to LTE or 5G networks wherein the MQTT server would reside within the cellular network, and its subscribers would be connected to the same network. 

\subsection{Deployment Diagram}

\begin{figure}[t]
\centering
{\includegraphics[clip,trim=0 0 0 0, width=\columnwidth]{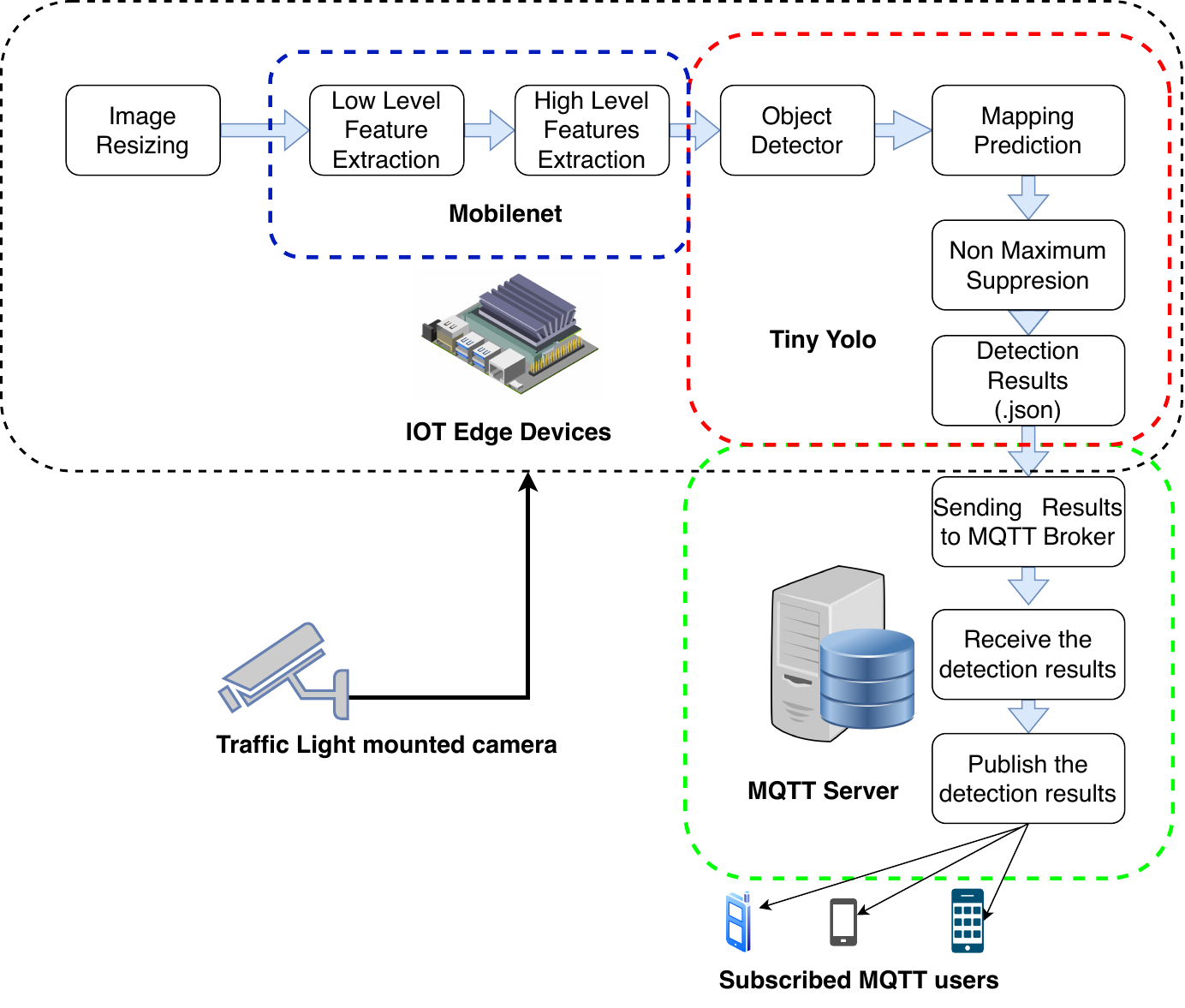}} 
\caption{Deployment diagram of the pedestrian detection system.  The traffic light-mounted camera transmits video to the AIoT Edge Device, which then pre-processes the data and runs the Tiny YOLO pedestrian detection model. The pedestrian detection results are sent to the MQTT Broker (external server), which then publishes the results to the subscribed MQTT users.}
\label{fig:System_Deployment}
\end{figure}
Fig. \ref{fig:System_Deployment} shows our system testbed topology. 
For DL processing, all input images must be resized to be similar to the images used for training. After that, the images are processed by three functional blocks: MobileNet, Tiny YOLO, and the MQTT server.

MobileNetV2 extracts feature maps using two stages. In the first stage, a convolutional layer captures low-level features such as edges, corners, and shapes. In the next stage, detection blocks attempt to capture high-level features. The detection blocks contain point-wise convolutions to project feature maps into a higher-dimensional space, allowing for the extraction of more complex patterns. Information is propagated from earlier to later layers, enabling the network to retain important low-level details while simultaneously capturing high-level features. Furthermore, MobileNetV2 performs down-sampling operations at specific points in the network. These operations reduce the spatial dimensions of the feature maps, effectively capturing features at different scales.

The YOLO model uses a Feature Pyramid Network (FPN) to improve the feature representation by generating a feature pyramid with multiple levels, each capturing features at different scales. This enables YOLOv3 to detect objects of varying sizes effectively. The FPN uses up-sampling and concatenation to combine features from different backbone levels. Lower-level features with higher spatial resolution are up-sampled and concatenated with higher-level features to create a multi-scale feature map. Then, features are concatenated from multiple levels. The features are then fed into subsequent layers for object detection using different scales and resolutions.
Finally, convolutional networks in the detection layers use concatenated feature maps to predict bounding boxes and class probabilities for detected objects at multiple scales. The detected objects, bounding box locations, and classification results are then converted to JSON format and sent to the MQTT server. 

When comparing video payload and JSON payload sizes, we find that the video payload is significantly larger, measuring 1500 kilobytes (KB) in size, while the JavaScript Object Notation (JSON) payload is much smaller, only 1200 bytes (B) in size. This significant disparity in size highlights the substantial difference in data between transmitting video content and JSON data. While the video payload requires efficient streaming protocols and specialized video delivery technologies to handle its payload size and real-time requirements, the JSON payload can be managed by lightweight protocols like MQTT, which is ideal for message queuing and delivery of smaller data packets. 





\section{Simulation and Analysis}
\label{sec:simresults}
We conducted simulations to evaluate the performance of our pedestrian detection model using two types of testing to showcase the model's capabilities. The first type of testing was conducted before deployment, aiming to assess the model's effectiveness in detecting pedestrians in the given context. We evaluated the model's performance using multiple metrics during this initial testing phase. The primary metric used for assessment was the Mean Average Precision (mAP), which measures the model's accuracy in detecting pedestrians. Additionally, we analyzed the model's computational efficiency by considering the number of Floating Point Operations Per Second (FLOPS) required to execute the DL model. Finally, we examined the model's size in terms of its parameters, which provides insights into the memory requirements and storage space. We used an RTX 3090 GPU with the Ubuntu 18.04 operating system to conduct these tests. The simulation used MMDetection version 2.25.1, Pytorch v.13.1, and Python 3.8. This configuration allowed us to thoroughly evaluate the model's performance and determine its suitability for pedestrian detection in our target scenario.

In the second testing stage, we deployed the pedestrian detection model on an edge AIoT device (Nvidia Jetson Nano). The objective of this phase was to evaluate the model's performance in real-time scenarios. Given the time-sensitive nature of pedestrian detection tasks, it was crucial for the model to accurately and swiftly identify pedestrians. We considered several key metrics to assess the model's capabilities. First, we examined the confidence score of the model, which represents its accuracy in correctly detecting pedestrians. Then, we analyzed the inference times, which indicate the speed at which the model can process and generate predictions. Finally, we evaluated the memory usage of the model to ensure it met the resource constraints of the edge AIoT device. We opted to deploy the model on a Jetson Nano device for this evaluation. The selection of the Jetson Nano was justified by its suitability for edge computing tasks, optimized hardware design for DL inference, and cost. The Jetson Nano balances computational power and energy efficiency, making it suitable for real-time pedestrian detection on edge devices. 

\subsection{Analysis}
\begin{table}[b]
\caption{Testing results using RTX 3090 GPU}
\label{tbl:Results_RTX3090}
\centering
\begin{tabular}{|l|l|l|l|}
\hline
\multicolumn{1}{|c|}{Model Name} & \multicolumn{1}{c|}{Accuracy} & \multicolumn{1}{c|}{FLOPS} & \multicolumn{1}{c|}{Parameters} \\ \hline
RetinaNet                        & 0.517                                       & 204.36 GFLOPs              & 36.1 M                          \\ \hline
Single Shot Detector             & 0.411                                       & 342.67 GFLOPs              & 24.39 M                         \\ \hline
TinyYOLOv3                       & 0.619                                       & \textbf{42.24 GFLOPs}               & \textbf{7.39 M}                          \\ \hline
Faster RCNN                      & 0.623                                       & 206.66 GFLOPs              & 41.12 M                         \\ \hline
EffficientNet                      & \textbf{0.6432}                                       &116.73 GFLOPs              & 18.34 M
                         \\ \hline
\end{tabular}
\end{table}

The testing was conducted using an RTX 3090 GPU and a subset of the Crowdhuman dataset, consisting of 4,370 annotated images.
A summary of the results obtained during the initial testing phase is provided in table \ref{tbl:Results_RTX3090}. 
It can be observed that the size of the model does not exhibit a linear relationship with its performance. Despite the Single Shot Detector model having the largest size and number of FLOPS, it performed poorly compared to other models. On the other hand, our model demonstrated compactness in terms of model size and FLOPS, but its mAP results were not the best. The EfficientNet model has the highest mAP, yet it requires over twice as many parameters as our model.
While these findings provide insights into our model's compactness and performance, it is important to note that further analysis is required to evaluate its real-time deployment capabilities. The speed and performance of the model in a real-time scenario need to be assessed to draw a comprehensive conclusion.

\begin{figure}[t]
\centering
\includegraphics[width=0.9\columnwidth]{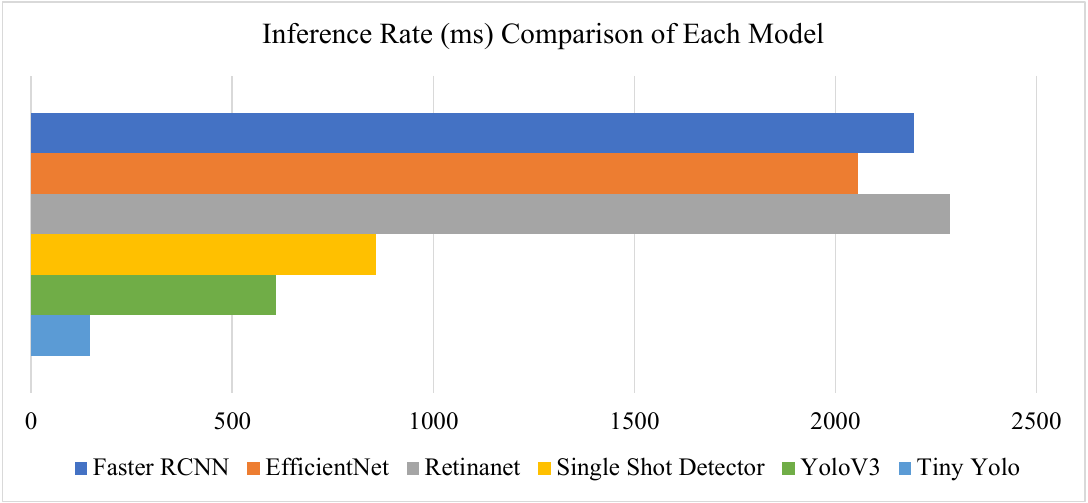}
\caption{Inference Time of each compared model (Jetson Nano)}
\label{fig:Infer_Rates}
\end{figure}
Fig. \ref{fig:Infer_Rates} shows that our model exhibits the highest inference rate / lowest inference time compared to the other models. It surpassed the slowest model by nearly $20\times$ and was over three times as fast as the second fastest model. This remarkable speed can be attributed to the optimization techniques that precisely reduce the convolutional layers. We achieved a more streamlined and efficient system by simplifying the model's architecture. Reducing convolutional layers led to faster inference times and contributed to a smaller model size. In the subsequent evaluation phase, we focus on assessing the accuracy of our optimized model.

\begin{figure}[t]
\centering
\includegraphics[width=0.9\columnwidth]{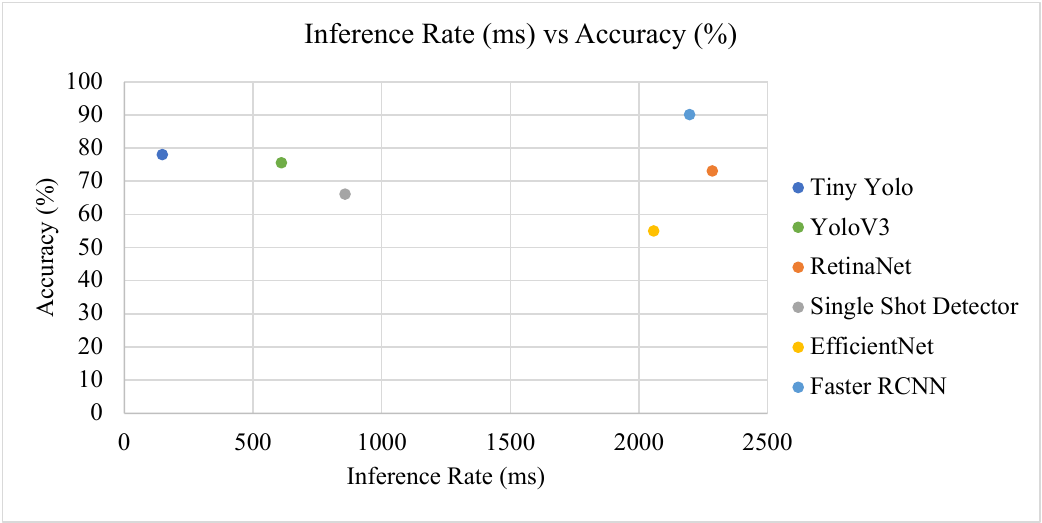}
\caption{Inference Time and accuracy of each model (Jetson Nano)}
\label{fig:Infer_Rate_vs_Acc}
\includegraphics[width=0.9\columnwidth]{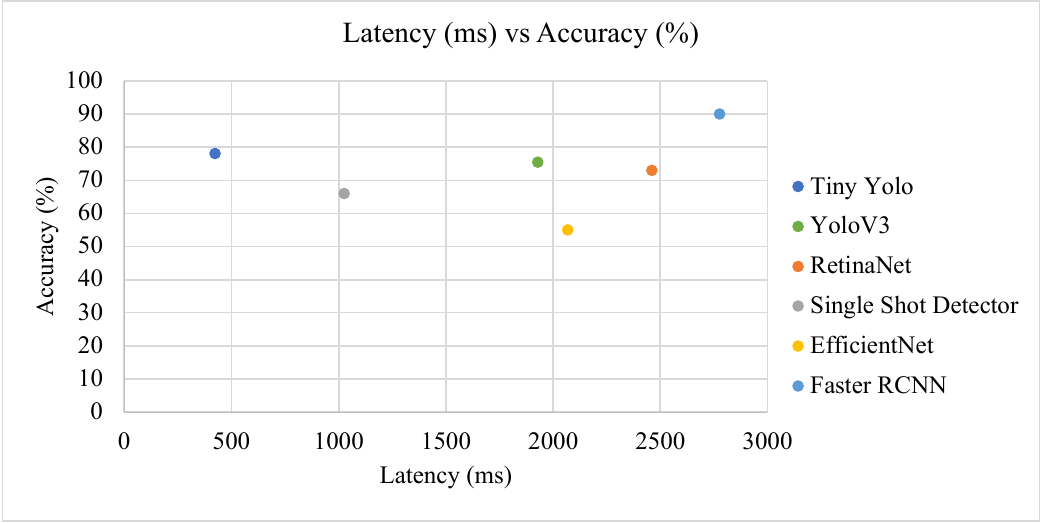}
\caption{Latency vs accuracy of each model (Jetson Nano)}
\label{fig:Latency_vs_Acc}
\includegraphics[width=0.9\columnwidth]{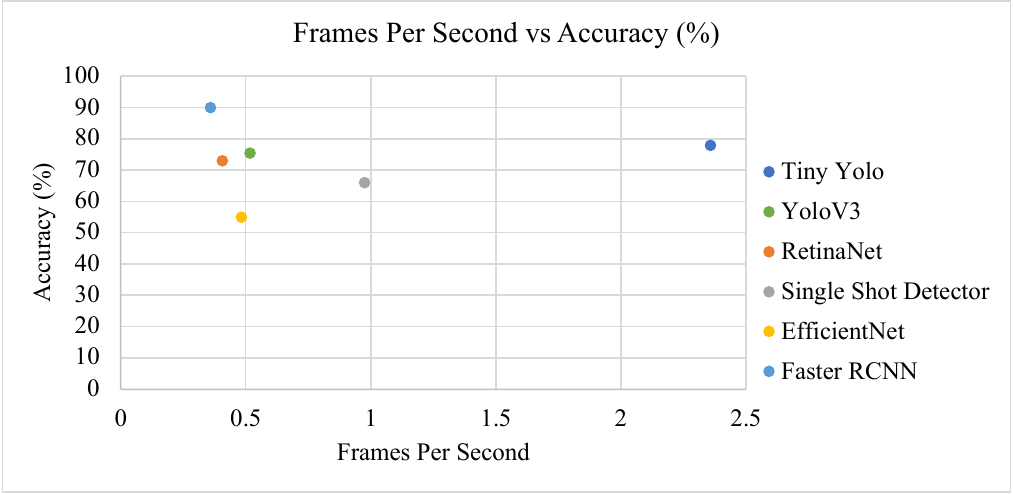}
\caption{Frames Per Second vs model accuracy (Jetson Nano)}
\label{fig:Acc_vs_Frame_Rate}
\end{figure}

Figs. \ref{fig:Infer_Rate_vs_Acc} illustrate that our model achieves a satisfactory accuracy of $78 \%$ while demonstrating significantly faster inference times. Although our model's accuracy is not the highest, it outperformed the Faster RCNN model in speed. In \cite{huang2017speed}, Huang et al. explained that in some cases, we have to decide whether to use a model with high accuracy or a fast inference time. 

In pedestrian detection, where real-time processing is crucial, a model with faster inference time is preferred. When considering pedestrian detection speed, we also consider the system model latency, pedestrian detection model inference time + time to transmit detection result to the MQTT server. This is captured by the system model latency shown in Fig. and \ref{fig:Latency_vs_Acc}. Based on the results, our system model is over twice as fast as the second-fastest model. Our proposed pedestrian detection system model has a latency of 424 ms; this is twice as fast as the second-fastest model. When considering the detection frame rate, as shown in Fig. \ref{fig:Acc_vs_Frame_Rate}, it becomes evident that the highest accuracy model (Faster RCNN) is unable to achieve sub-second pedestrian detection, i.e., it cannot effectively detect pedestrians in real-time (frame rate $\ge 1$) scenarios.

\begin{figure}[t]
\centering
\includegraphics[width=0.9\columnwidth]{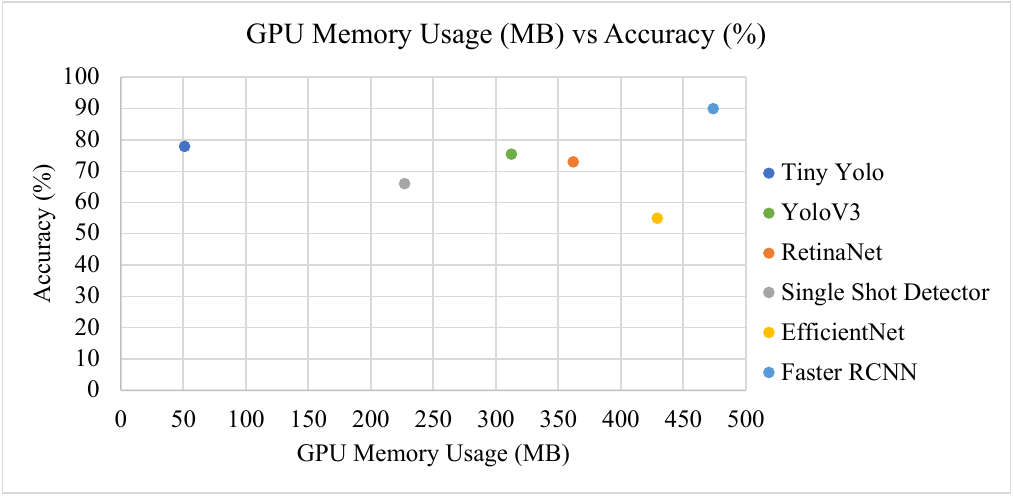}
\caption{Accuracy and memory usage of each model (Jetson Nano)}
\label{fig:Infer_Rate_vs_GPU_memory}
\end{figure}

Fig. \ref{fig:Infer_Rate_vs_GPU_memory} presents the model evaluation results, indicating that our optimized model requires significantly less memory than larger models. Despite its compact size, our model performs similarly to the larger models. Specifically, our model requires only $~10\%$ of the resources used by the largest model, Faster RCNN, while achieving a similar performance. This highlights the efficiency and resource-saving benefits of our optimized model.

\begin{table}[b]
\caption{Comparison of Tiny YOLO and YOLO (Jetson Nano)}
\label{tbl:Tiny_YOLO_vs_Orig_YOLO}
\centering
\begin{tabular}{|p{4em}|l|p{5em}|p{4em}|p{5em}|p{4em}|}
\hline
Name & Accuracy & FLOPS & Param & GPU Mem (MB) & Inference (ms)\\ 
\hline
Tiny YOLOv3 & \textbf{0.78} & \textbf{42.24 GFLOPs} & \textbf{7.39 M} & \textbf{50.81} & \textbf{423.73} \\ 
\hline
YOLOv3 & 0.755 & 193.85 GFLOPs & 61.52 M & 312.48 & 610.13 \\ 
\hline
\end{tabular}
\end{table}

We now compare our optimized YOLOv3 model to the original YOLOv3 model 
to determine if the optimizations we applied to our model improved or harmed the model's performance.
Based on the comparison of both the Tiny YOLO and the YOLO models, see table \ref{tbl:Tiny_YOLO_vs_Orig_YOLO}, it is clear that our optimization results in better performance and a lower computational power requirement. Although the accuracy improvement is minor (only $2.5\%$), other results, such as GPU memory and inference times, improve significantly. Our model used $6\times$ less GPU memory and achieved around $5\times$ faster inference rate compared to the vanilla YOLOv3 model.

A primary goal in computer vision is to improve accuracy. However, this goal is hampered by achieving higher accuracy, which frequently necessitates using larger and more complex models. When designed for deployment on resource-constrained platforms, such as low-resource devices common in the AIoT domain, such models are presented with inherent challenges, such as limited memory, processing power, and energy. 
As a result, while a larger model may outperform others in accuracy, its viability for real-world application becomes questionable due to its incompatibility with the need for real-time resource-efficient processing.

\section{Conclusion and Future Work}
\label{sec:conclusion}
Based on our research, deploying the optimized DL model on IoT edge devices has yielded excellent real-time pedestrian detection results. The model's efficient optimization allows it to run on resource-constrained devices with minimal computational power, achieving high accuracy and fast detection. 


Integrating our system with commercially available LTE or 5G networks in the future presents a significant challenge. This would require ensuring seamless compatibility with cellular networks to enhance system performance. Ultimately, our system provides an affordable and intelligent alternative for intelligent transportation systems, promising significant benefits in cost-effective and efficient solutions.

\FloatBarrier

\bibliography{IEEEabrv, my_bib_IEEE}
\bibliographystyle{IEEEtran}

\begin{IEEEbiography}[{\includegraphics[width=1in,height=1.25in,clip,keepaspectratio]{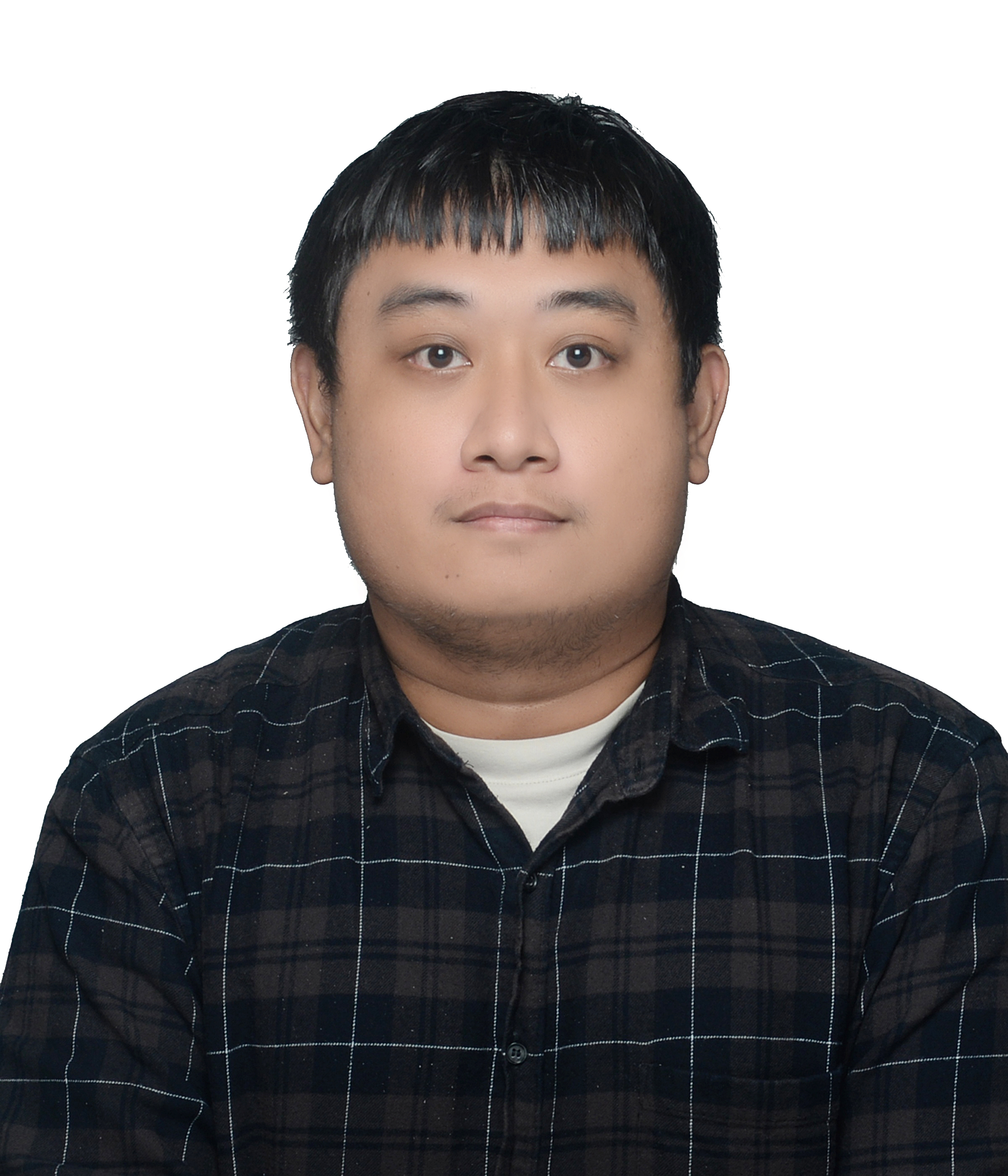}}]{Muhammad Dany Alfriki}
graduated from the Department of Computer Science and Information Engineering, National Taiwan University of Science and Technology, Taipei, Republic of China, in 2023. Currently he is working as a Software Engineer, his research interests include computer vision, embedded systems, and large language model.
\end{IEEEbiography}

\begin{IEEEbiography}[{\includegraphics[width=1in,height=1.25in,clip,keepaspectratio]{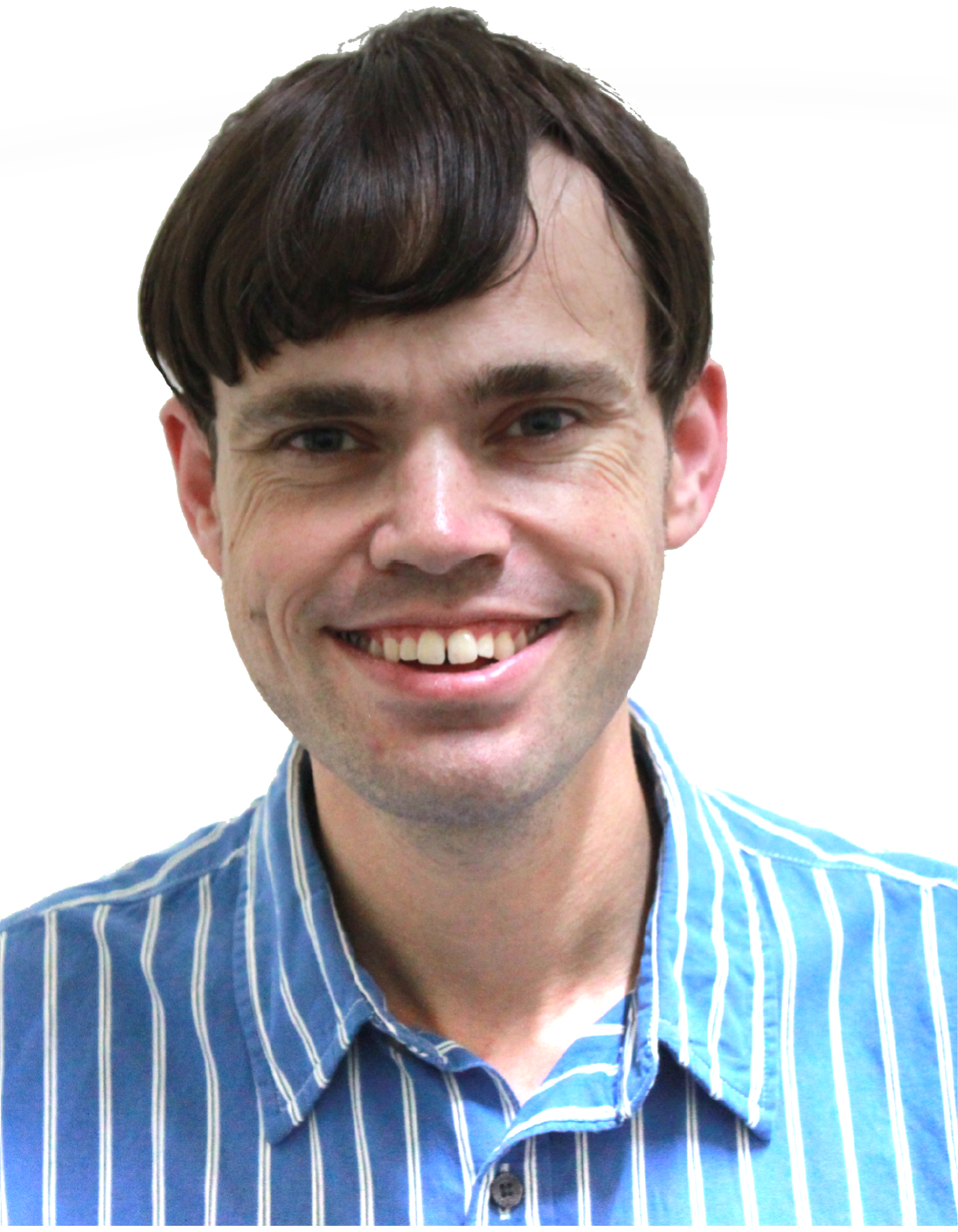}}]{Rafael Kaliski}
obtained his PhD from National Taiwan University in 2017.  He is currently an assistant professor in the department of Computer Science and Engineering at National Sun Yat-sen University (Taiwan).  His research interests include wireless networks, multimedia, resource allocation, cyber security, and optimization (machine learning, game theory, and mathematical programming).
\end{IEEEbiography}

\end{document}